\documentclass[accepted]{article}

\usepackage{microtype}
\usepackage{graphicx}
\usepackage{subcaption}
\usepackage{booktabs}
\usepackage{amssymb}
\usepackage{multirow}
\usepackage{amsmath}
\usepackage{rotating}

\usepackage{hyperref}

\usepackage[accepted]{icml2021}

\icmltitlerunning{ Learning a powerful SVM using piece-wise linear loss functions}

\begin{document}

\twocolumn[
\icmltitle{Learning a powerful SVM using piece-wise linear loss functions}



\icmlsetsymbol{equal}{*}

\begin{icmlauthorlist}
\icmlauthor{Pritam Anand}{to}
\\ $^1$ Department of Computer Science, South Asian University, New Delhi-110021.\\
Email :- ltpritamanand@gmail.com
\end{icmlauthorlist}

\icmlaffiliation{to}{Department of Computer Science, South Asian University,  New Delhi, India}

\icmlcorrespondingauthor{Pritam Anand}{ltpritamanand@gmail.com}

\icmlkeywords{Machine Learning, ICML}

\vskip 0.3in
]




\begin{abstract}
	In this paper, we have considered general $k$-piece-wise linear convex loss functions in SVM model for measuring the empirical risk. The resulting  $k$-Piece-wise Linear loss Support Vector Machine ($k$-PL-SVM) model is an adaptive SVM model which can learn a suitable piece-wise linear loss function according to nature of the given training set.  The $k$-PL-SVM models are general SVM models and existing popular SVM models, like C-SVM, LS-SVM and Pin-SVM models, are their particular cases. We have performed the extensive numerical experiments with $k$-PL-SVM models for $k$ = 2  and 3 and shown that they are improvement over existing SVM models.
\end{abstract}

Support Vector Machine (SVM) models \cite{VapnikCortes}\cite{statistical_learning_theory}\cite{GUNNSVM} are still very useful and popular among researchers. It is because of their interesting characteristics which remain missing in other machine learning models. SVM models implement the Structural Risk Minimization (SRM) principle \cite{statistical_learning_theory} and can explicitly minimize the regularization in its optimization problem to avoid over-fitting. Most of the existing SVM models require to solve appropriate convex programming problems only which guarantees the global optimal solution. Further, there are different choices of the specific loss functions and kernel functions \cite{mercer1909xvi} available which can be used in SVM model according to the characteristics of the given dataset.

 For a given binary classification problem with the training set  $ T= \{(x_i,y_i):x_i \in \mathbb{R}^n, y_i\in \{ -1,1\}, i=1.2,...,\textit{l}  \}$, the SVM model obtains the  kernel generated decision function $sign(w^T\phi(x)+b)$. For this, the SVM models solve the optimization problem in which good trade-off between the empirical error of the training set and the regularization is minimized efficiently.

In SVM models, we use the loss function to measure the empirical risk of given training set. The characteristics and performance of a SVM model depends upon the way it measures the empirical error of the given training set. Therefore,  the choice of loss function is very crucial in SVM models.

The standard C-SVM model uses the Hinge loss function to measure the empirical risk. The Hinge loss function is given by $ L_{Hinge}(u) = max(u,0),~ u \in \mathbb{R}$. For training set  $T$, the C-SVM model solves the optimization problem
\begin{equation}
\min_{(w,b)}  \frac{1}{2} ||w||_2^2+ C_0\sum_{i=1}^{l} L_{Hinge}(1-y_i(w^T\phi(x_i)+b)),
\label{CSVM}
\end{equation}
where $C_0$ is an user supplied parameter which is used for tuning the trade-off between the empirical error and the model complexity. The use of the Hinge loss function in C-SVM model let us obtain its geometrical interpretation. The solution of C-SVM model is geometrically equivalent to obtaining a separating hyperplane $(w^T\phi(x)+b)=0$ in the feature space with maximum margin.

\begin{figure*}[h]
	\centering
	\subfloat[Hinge Loss ]{\includegraphics[width=.25\linewidth, height = 0.12\textheight ]{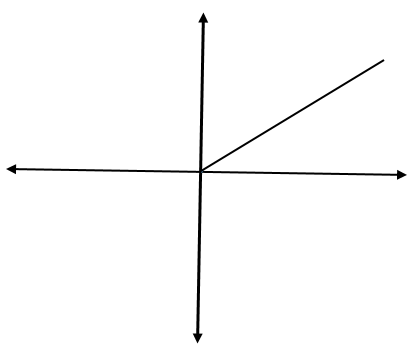}}
	\subfloat[Pinball loss]{\includegraphics[width=.25\linewidth,height = 0.12\textheight]{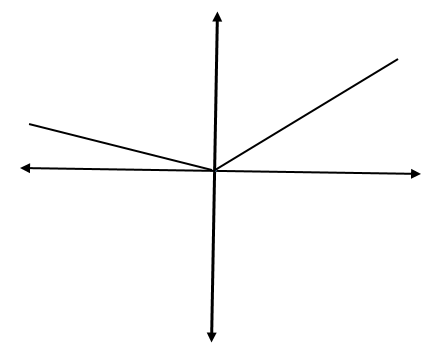}}
	\subfloat[ Least squares loss ]{\includegraphics[width=.25\linewidth,height = 0.12\textheight]{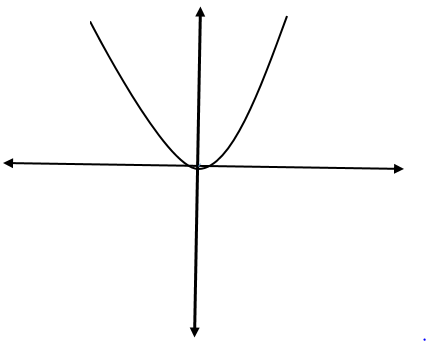}}
	\caption{Existing loss functions in SVM}
\end{figure*}

The Least Squares Support Vector Machine (LS-SVM) model \cite{LSSVR2} uses the well known least squares loss function \cite{legendre1805nouvelles} to measure the empirical error, which is given by $ L_{}(u) = u^2,~ u \in \mathbb{R}$. For the given training set $T$, the LS-SVM model solves the optimization problem
\begin{equation}
\min_{(w,b)}  \frac{1}{2} ||w||_2^2+ C_0\sum_{i=1}^{l} (1-y_i(w^T\phi(x_i)+b))^2.
\end{equation}

The Pin-SVM model \cite{pinsvm} minimizes the pinball loss function \cite{quantile1}\cite{pinsvm} to measure the empirical error of the training set. For $-1 \leq \tau \leq 1$, it is given by $L_{\tau}(u) = max (u, -\tau u),~u \in \mathbb{R}$. For the given training set, it solves the optimization problem 
\begin{equation}
\min_{(w,b)}  \frac{1}{2} ||w||_2^2+ C_0\sum_{i=1}^{l} L_{\tau}(1-y_i(w^T\phi(x_i)+b)). \label{pinsvm}
\end{equation}

We have plotted the Hinge loss function, the Least squares loss function and the pin-ball loss function in Figure 1. We can realize that these loss functions based SVM models, particularity the C-SVM  and LS-SVM models,  are rigid in nature. They measure the empirical error of the given training set without bothering the nature of data. The loss functions used in these SVM models do not posses the capability of adapting themselves according to the nature of the given data. Therefore, there is  the need of introducing adaptive and flexible loss functions in SVM models.

We also note that the linear loss functions are robust. The influence of any outlier data point on the decision function obtained by the linear loss functions is limited. But, the least squares loss function and other polynomial loss functions are not robust.

In this paper, we have introduced  a family of  $k$- piece-wise linear loss functions and used them in SVM model. The  $k$- piece-wise linear loss functions are convex and robust. The use of these loss functions in SVM results into an adaptive SVM model which can learn a suitable piece-wise linear loss function according to the nature of the data. The proposed $k$-piece-wise Linear loss function based Support Vector Machine ($k$-PL-SVM) model is a general SVM model. Most of popular SVM models, like C-SVM, Pin-SVM and LS-SVM  models, are its particular cases. We have briefly described the other interesting characteristics of the proposed $k$-PL-SVM in this paper. Further, we have carried out an extensive numerical experiments with $k$-PL-SVM for $k =$ 2 and 3. These results conclude that the proposed $k$-PL-SVM model can obtain significant improvement in prediction over existing SVM models.

We have organized the rest of this paper as follows. We have described our $k$-piece-wise linear loss functions and  the resulting SVM model in Section-2 of this paper,. Section-3 describes some interesting properties of the proposed $k$-PL-SVM model. In section-4, we have presented the extensive numerical results to realize that the proposed $k$-PL-SVM model has a lot of potential for improving the prediction of SVM models. Section-5 concludes this paper.
\vspace{-0.1 in}
\section{Piece-wise linear loss function based Support Vector Machine}
We  propose the general piece-wise linear convex loss function  for SVM model. The k-piece-wise linear loss function is defined as
{\small\begin{eqnarray}
L^k(u)=max( u,-\tau_1u+\epsilon_1,-\tau_2u+\epsilon_2 ,....,-\tau_{k-1}u+\epsilon_{k-1}), \nonumber  \hspace{-10mm}\\ u \in \mathbb{R} \label{pllos}
\end{eqnarray}}
where $\tau_1$,$\tau_2$,...,$\tau_{k-1}$ and  $\epsilon_1$,$\epsilon_2$,...,$\epsilon_{k-1}$ are real valued parameters.\\ \\ 
\textbf{Preposition 1:-} Let $F^{(k)}$ be the collection of all k-piece-wise linear convex functions. Let, $f^{(k)}\in F^{(k)}$ be such that $f^{(k)}(u) =u$,  $\forall u \in [ u_0, u'_0]$, for some $u_0$, $u'_0 > 0$. Then, the  loss function $L^k(u)$ is sufficient to represent the family $F^{(k)}$.\\ \\
\textbf{Proof:-} We shall prove the statement by using the principle of induction.

At first, we shall show that the elements of $F^{(2)}$ can be obtained by our proposed loss function $L^2(u) = max( u, -\tau_1u + \epsilon_1$). One of the segmented line of an arbitrary $f^{(2)} \in F^{(2)}$  would be  $y= u$ as it must satisfy $f(u) =u$,$\forall u \in [ u_0, u'_0]$ for some $u_0$ and $u'_0 > 0$ .   Let  the another segmented line of the  $f^{(2)}$ is $ y= au+b$ then for $\tau_1= -a$ and $\epsilon_1 = b$ , the $f^{(2)}$ can be represented by 
the $ L^2(u) = max( u, -\tau_1u+\epsilon_1)$.

Further, let $F^{(m)}$ can be represented by the loss function $L^m(u)$. It means that $f^{(m)}(u)$ can be obtained from $max( u, -\tau_1u + \epsilon_1,...., -\tau_{m-1}u + \epsilon_{m-1}) $. Then we need to show  that  $f^{(m+1)}$ can be obtained by the loss function $L^{m+1}$ for completing the proof. The $f^{(m+1)}$ can be constructed by  considering an additional segmented line $ y= a'u+b'$. Since $f^{(m+1)}$ has to  remain convex, so it can be obtained by $max(L^m(u), -\tau_{m}u+\epsilon_{m} )$ with $\tau_{m} = -a' $ and $\epsilon_{m}= b$. It means that the $f^{(m+1)}$ can be obtained from  $ L^{m+1}(u) = max( u, -\tau_1u + \epsilon_1,...., -\tau_{m}u + \epsilon_{m}). $

$~~~~~~~~~~~~~~~~~~~~~~~~~~~~~~~~~~~~~~~~~~~~~~~~~~~~~~~~~~~~~~~~~~~~~~~~~~~~~~~~~~~~~~~~~~~~~~~~~~~~~~~~~~~~~~~~~~~~~~~~~~~~~~~~~~~~~~~~~~~~~~~~~\square$

\begin{figure}[h]
	\centering
	\subfloat[]{\includegraphics[width=.25 	\linewidth]{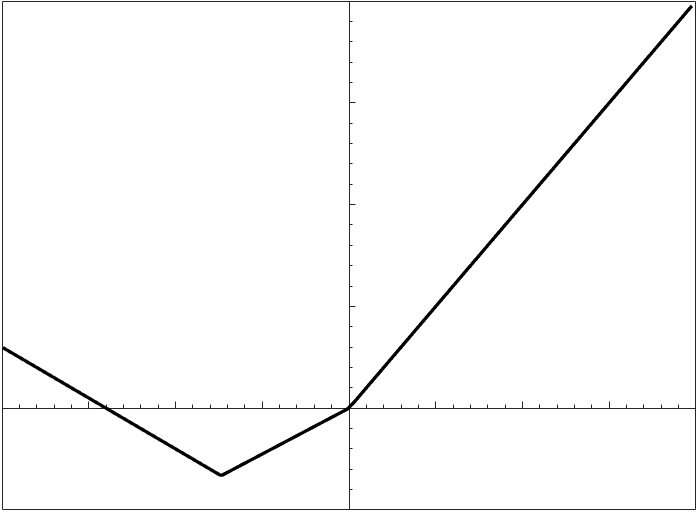}}
	\subfloat[]{\includegraphics[width=.25\linewidth]{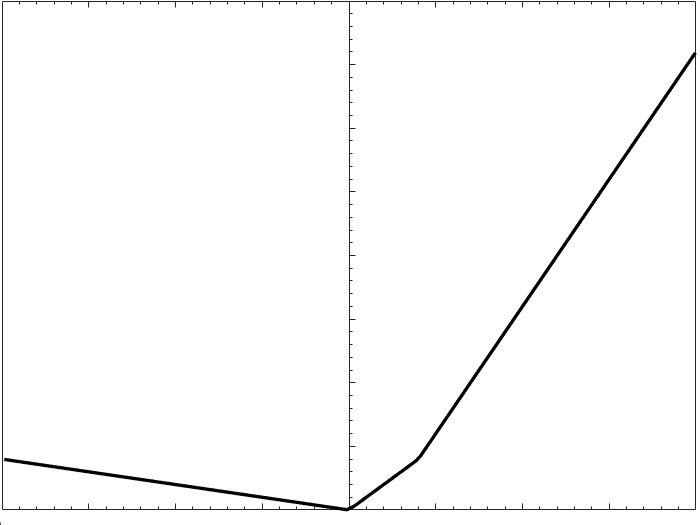}}
	\subfloat[]{\includegraphics[width=.25\linewidth]{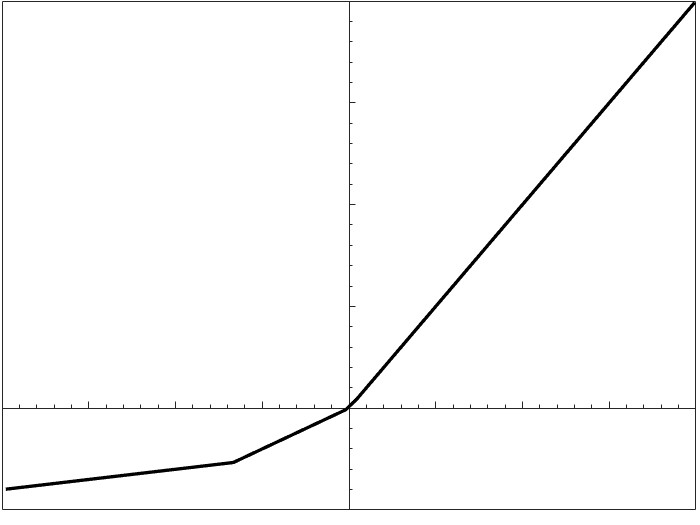}}\\
	\subfloat[]{\includegraphics[width=.25\linewidth]{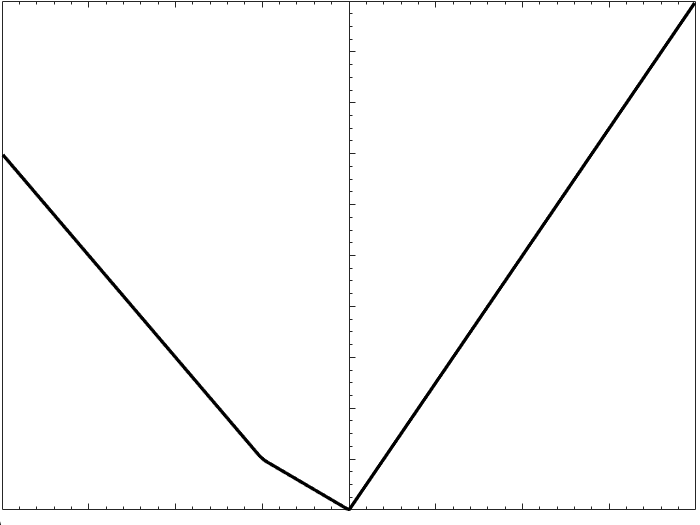}}
	\subfloat[]{\includegraphics[width=.25\linewidth]{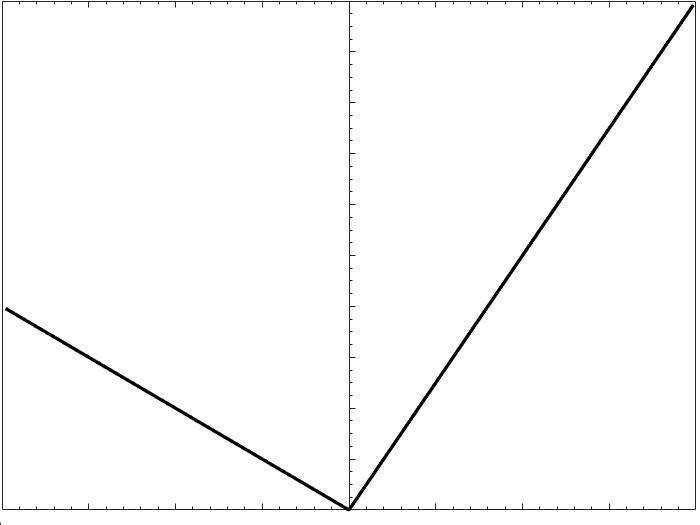}}
	\subfloat[]{\includegraphics[width=.25\linewidth]{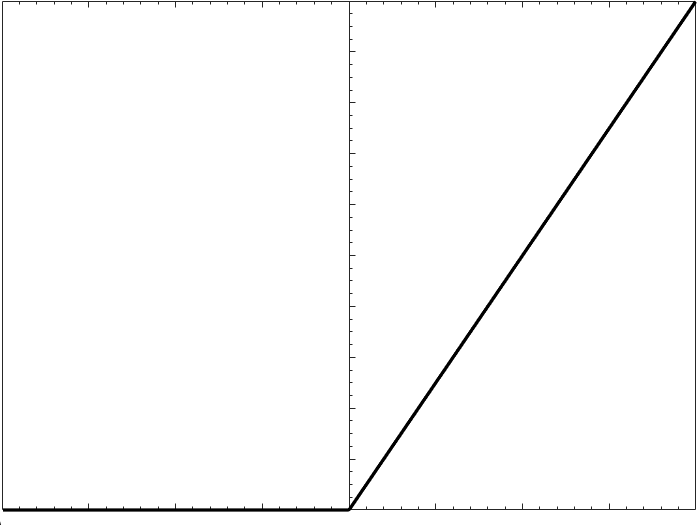}}
	\caption{The 3-piece-wise linear loss function for (a) $\tau_1= -0.45 ,\tau_2= 0.50, \epsilon_1=0 , \epsilon_2=-7$ (b) $\tau_1= -2 ,\tau_2= 0.2, \epsilon_1=-4 , \epsilon_2= 0$ (c)  $\tau_1= -0.4 ,\tau_2= -0.1, \epsilon_1=0 , \epsilon_2= -2$ (d)  $\tau_1= 0.4 ,\tau_2= 0.8, \epsilon_1=0 , \epsilon_2= -2$ (e)  $\tau_1= 0.4 ,\tau_2= 0, \epsilon_1=0 , \epsilon_2= 0$ (f)  $\tau_1= 0 ,\tau_2= 0, \epsilon_1=0 , \epsilon_2= 0$.}
	\label{steady_state}
\end{figure}

Figure \ref{steady_state} shows the $3$-piece-wise linear loss function for some particulars values of parameters. It can be noted that it can also reduce to the pinball loss function and Hinge loss function for the particular chosen values of its parameters.

For a given classification task with training set $ T~=~\{{(x_{i},y_{i}})|x_{i}~\in~\mathbb{R}^n,~y_{i} \in  \{1,-1\},~i=1,2,.,l \}$,  we used the proposed $k$-piece-wise linear loss function for measuring the empirical risk.  The resulting $k$-Piece-wise Linear loss Support Vector Machine ($k$-PL-SVM) model minimizes the empirical risk obtained by the chosen $k$-piece-wise linear loss function along with the regularization term $\frac{1}{2}||w||_2$  in its optimization problem as follow.
\begin{eqnarray}
\min_{(w,b)} \frac{1}{2}||w||_2 + C\sum_{i=1}^{l}L^{k}(1-y_i(w^T\phi(x_i)+b))    \nonumber\\
&\hspace{-75mm} = \min\limits_{(w,b)} \frac{1}{2}||w||_2 + C\sum\limits_{i=1}^{l} max \bigg( 1-y_i(w^T\phi(x_i)+b)),  \nonumber\\ &\hspace{-73mm}-\tau_1(1-y_i(w^T\phi(x_i)+b))+\epsilon_1,...,-\tau_{k-1}(1  -y_i\nonumber\\ &\hspace{-105mm}(w^T\phi(x_i)  +b))    +\epsilon_{k-1}. \bigg) 
\label{op_PLSVM}
\end{eqnarray}
Let us consider the $l$-dimensional slack variable $\xi$ such that $\xi_i =  max((1-y_i(w^T\phi(x_i)+b)),-\tau_1(1-y_i(w^T\phi(x_i)+b))+\epsilon_1,...,-\tau_{k-1}(1-y_i(w^T\phi(x_i)+b))+\epsilon_{k-1})$ then the optimization problem (\ref{op_PLSVM}) can be converted to the following Quadratic Programming Problem (QPP)
\begin{eqnarray}
& \hspace{-50mm}\min\limits_{(w,b,\xi)}~~ \frac{1}{2}||w||_2 + C_0\sum\limits_{i=1}^{l} \xi_i \nonumber \\
& \hspace{-70mm}\mbox{subject to, }  \nonumber \\
& \hspace{-45mm} \xi_i \geq 1-y_i(w^T\phi(x_i)+b), \nonumber \\
& \hspace{-30mm} \xi_i \geq -\tau_1(1-y_i(w^T\phi(x_i)+b)+ \epsilon_1, \nonumber \\ 
& \hspace{-30mm} ................................................., \nonumber \\ 
& \hspace{-30mm} ................................................., \nonumber \\ 
& \hspace{-5mm} \xi_i \geq -\tau_{k-1}(1-y_i(w^T\phi(x_i)+b)+ \epsilon_{k-1}, i=1,2,..,l~
\label{pllsvm}
\end{eqnarray}
where  $\tau_1 , \tau_{2},....,\tau_{k-1}, \epsilon_1, \epsilon_2,.....\epsilon_{k-1}$ and $C_0 \geq 0 $ are user supplied parameters. The parameter $C_0$ can be used to control the trade-off between the empirical error and model complexity. To handle the unbalanced class labeling problem , we may consider a $l$-dimensional vector $C=(C_1,C_2,\ldots C_l)$   in the place of single constant $C_0$  such that 
\begin{equation}
C_i= 
\begin{cases}
C_0,~~~~~~~~~~~~  y_i = + 1 ,\\
p C_0 ,~~~~~~~~~~~~ y_i = -1,
\end{cases} \label{ci}
\end{equation}
where $p$ is defined as $p=~\frac{\mbox{number of data points on `class +1'}}{\mbox{number of data points in `class -1'}}$.  Thereafter, we have preferred to solve the following optimization problem for our  $k$-PL-SVM model. 
\begin{eqnarray}
& \hspace{-50mm}\min\limits_{(w,b,\xi)}~~ \frac{1}{2}||w||_2 + \sum\limits_{i=1}^{l}C_i \xi_i \nonumber \\
& \hspace{-69mm}\mbox{subject to, }  \nonumber \\
& \hspace{-47mm} \xi_i \geq 1-y_i(w^T\phi(x_i)+b), \nonumber \\
& \hspace{-32mm} \xi_i \geq -\tau_1(1-y_i(w^T\phi(x_i)+b))+ \epsilon_1, \nonumber \\
& \hspace{-32mm} \xi_i \geq -\tau_2(1-y_i(w^T\phi(x_i)+b))+ \epsilon_2, \nonumber \\
& \hspace{-32mm} ................................................., \nonumber \\ 
& \hspace{-32mm} ................................................., \nonumber \\ 
& \hspace{-8mm} \xi_i \geq -\tau_{k-1}(1-y_i(w^T\phi(x_i)+b)+ \epsilon_{k-1}, i=1,2,....l.
\label{pll_primal}
\end{eqnarray}
To derive the Wolfe dual, we need to obtain the Lagrangian function for the primal problem (\ref{pll_primal})  of our  $k$-PL-SVM model. The Lagrangian function can be obtained as

$ L (w,b,\xi, \alpha ,\alpha^{(1)} ,\alpha^{(2)} , ...,\alpha^{(k-1)})  = \frac{1}{2}||w||_2 + C_0\sum\limits_{i=1}^{l} \xi_i 
  - \sum\limits_{i=1}^{l} \alpha_i( y_i(w^T\phi(x_i)+b) -1+ \xi_i) 
  -\sum \limits_{i=1}^{l} \alpha^{(1)}_i(\tau_1(1-y_i(w^T\phi(x_i)+b)) + \xi_i -\epsilon_1) -\sum \limits_{i=1}^{l} \alpha^{(2)}_i(\tau_2(1-y_i(w^T\phi(x_i)+b)) + \xi_i -\epsilon_2) ~- ~... 
.....-\sum \limits_{i=1}^{l} \alpha^{(k-1)}_i(\tau_{k-1}(1-y_i(w^T\phi(x_i)+b)) + \xi_i -\epsilon_{k-1}).$\\
Here $\alpha, \alpha^{(1)},\alpha^{(2)},...,\alpha^{(k-1)} > 0$ are $l$-dimensional vectors of Lagrangian multipliers. We list the Karush-Kuhn-Tucker (KKT) conditions for the primal problem (\ref{pll_primal}) as follow. 
\begin{footnotesize}
	\begin{align}
  &  w= \sum\limits_{i=1}^{l}(\alpha_i-\tau_1\alpha^{(1)}_i -....-\tau_{k-1}\alpha^{(k-1)}_i )y_i\phi(x_i) , \\	
  & \sum\limits_{i=1}^{l}(\alpha_i-\tau_1\alpha^{(1)}_i -....-\tau_{k-1}\alpha^{(k-1)}_i )y_i= 0,  \\
 & C_i- \alpha_i - \alpha^{(1)}_i- ....-\alpha^{(k-1)}_i = 0, ~ i=1,2,..l,   \\
 & {\alpha_i}( y_i(w^T\phi(x_i)+ b)-1+\xi_i) = 0, i=1,2,...,l,\label{ur2}\\
 & {\alpha^{(m)}_i}( \tau_m(1-y_i(w^T\phi(x_i)+ b))+\xi_i-\epsilon_m)=0,~ \nonumber \\ &  i=1,2,..l,~m=1,2,,...k-1, \label{ur5}\\
 & \xi_i \geq 1- y_i(w^T\phi(x_i)+b),~ i=1,2,..l,\label{ur8}\\
 & \xi_i \geq -\tau_m(1- y_i(w^T\phi(x_i)+b)) + \epsilon_{m}, \nonumber \\ &  ~i=1,2,..l, ~m=1,2,,..,k-1,~~~\label{ur9} \\       
  & \alpha_{i}\geq 0 , ~ i=1,2,..,l,~      \nonumber \\ 
 &  \alpha^{(m)}_i\geq 0 , ~ i=1,2,..,l,~   ~m=1,2,,...k-1.       \label{ur7} 
\end{align}
\end{footnotesize}
Using the above KKT conditions, the Wolfe dual of the primal problem (\ref{pll_primal})  of our  $k$-PL-SVM model can be obtained as
{\small \begin{eqnarray}
\min_{(\alpha,\alpha^{(1)},...,\alpha^{(k-1)})} \frac{1}{2}\sum_{j=1}^{l}\sum_{i=1}^{l}(\alpha_{i}-\tau_1\alpha^{(1)}_{i}-,..~,-\tau_{k-1}\alpha^{(k-1)}_{i})  \nonumber \\   
& \hspace{-90mm} y_iy_j\phi(x_i)^T\phi(x_j)(\alpha_{j}-\tau_1\alpha^{(1)}_{j}-,..~,-\tau_{k-1}\alpha^{(k-1)}_{j})    \nonumber \\
  & \hspace{-90mm}- \sum\limits_{i=1}^{l}(\alpha_{i}-\tau_1\alpha^{(1)}_{i}-..~,-\tau_{k-1}\alpha^{(k-1)}_{i})  
  - \sum\limits _{i=1}^{l}(\alpha^{(1)}_{i}\epsilon_1 +,.. \nonumber \\
 & \hspace{-130mm}  ..,+\alpha^{(k-1)}_{i}\epsilon_{k-1} )\nonumber \\
& \hspace{-150mm}\mbox{subject to,} \nonumber \\
& \hspace{-80mm} \sum\limits_{i=1}^{l}(\alpha_{i}-\tau_1\alpha^{(1)}_{i}-,..~,-\tau_{k-1}\alpha^{(k-1)}_{i})y_i=0, \nonumber \\
& \hspace{-75mm} C_i- \alpha_i - \alpha^{(1)}_i-,...,-\alpha^{(k-1)}_i = 0,~ i=1,2,..l, \nonumber \\
& \hspace{-76mm} \alpha_{i}\geq 0 , ~\alpha^{(m)}_i\geq 0 ,~ i=1,2,..,l,~ ~m=1,2,,...k-1.
\end{eqnarray}
}For a given positive semi-definite kernel $k$, satisfying  Mercer condition (Mercer,\cite{mercer1909xvi}), we can obtain  $k(x_i,x_j)= \phi(x_i)^T\phi(x_j)$ without explicit knowledge of mapping $\phi$. It makes the above dual problem  to reduce as 
{\small \begin{eqnarray}
	\min_{(\alpha,\alpha^{(1)},...,\alpha^{(k-1)})} \frac{1}{2}\sum_{j=1}^{l}\sum_{i=1}^{l}(\alpha_{i}-\tau_1\alpha^{(1)}_{i}-,..~,-\tau_{k-1}\alpha^{(k-1)}_{i})  \nonumber \\   
	& \hspace{-90mm} y_iy_jk(x_i,x_j)(\alpha_{j}-\tau_1\alpha^{(1)}_{j}-,..~,-\tau_{k-1}\alpha^{(k-1)}_{j})    \nonumber \\
	& \hspace{-90mm}- \sum\limits_{i=1}^{l}(\alpha_{i}-\tau_1\alpha^{(1)}_{i}-..~,-\tau_{k-1}\alpha^{(k-1)}_{i})  
	- \sum\limits _{i=1}^{l}(\alpha^{(1)}_{i}\epsilon_1 +,.. \nonumber \\
	& \hspace{-130mm}  ..,+\alpha^{(k-1)}_{i}\epsilon_{k-1} )\nonumber \\
	& \hspace{-150mm}\mbox{subject to,} \nonumber \\
	& \hspace{-80mm} \sum\limits_{i=1}^{l}(\alpha_{i}-\tau_1\alpha^{(1)}_{i}-,..~,-\tau_{k-1}\alpha^{(k-1)}_{i})y_i=0, \nonumber \\
	& \hspace{-75mm} C_i- \alpha_i - \alpha^{(1)}_i-,...,-\alpha^{(k-1)}_i = 0,~ i=1,2,..l, \nonumber \\
	& \hspace{-76mm} \alpha_{i}\geq 0 , ~\alpha^{(m)}_i\geq 0 ,~ i=1,2,..,l,~ ~m=1,2,,...k-1. \label{dual_plsvm}
	\end{eqnarray}
}
After obtaining the solution vectors $\alpha,\alpha^{(1)},...,\alpha^{(k-1)}$ of the dual problem (\ref{dual_plsvm}), we can classify an unseen data point $ x \in \mathbb{R}^n$ using the decision function
\[f(x)= sign(w^T\phi(x) +b ) \] \[= sign (~ \sum_{i=1}^{l}(\alpha_{i}-\tau_1\alpha^{(1)}_{i}-,..~,-\tau_{k-1}\alpha^{(k-1)}_{i})y_ik(x_i,x) +b ~). \]

\paragraph{Obtaining the value of b:- }
For $\alpha_{j} > 0$, $\alpha^{(m)}_{i} > 0$ and $\tau_m \ne -1$, $m=1,2,..,k-1$, we can obtain using the KKT conditions (\ref{ur2}) and (\ref{ur5})
\[  y_j(w^T\phi(x_j)+ b)-1+\xi_j =0 \] ~~ and ~~
\[\tau_m(1-y_j(w^T\phi(x_j)+ b))+\xi_j-\epsilon_m=0,\]   which gives

\[ b= y_j- \bigg(\sum_{i=1}^{l} (\alpha_{i}-\tau_1\alpha^{(1)}_{i}-,..~,-\tau_{k-1}\alpha^{(k-1)}_{i})y_ik(x_i,x_j)\] \[ \hspace{-30mm}- y_j\frac{\epsilon_{m}}{(1+\tau_m)} \bigg).\]

Also for  $\alpha^{(m_1)}_{j} >0$ , $\alpha^{(m_2)}_{j} >0$ , $m_1 \ne m_2$ and $\tau_{m_1} \ne \tau_{m_2}$, $m_1,m_2=1,2,..,k-1$, we can obtain using the KKT conditions (\ref{ur5})
\[\tau_{m1}(1-y_j(w^T\phi(x_j)+ b))+\xi_j-\epsilon_{m1}=0,\] ~ and ~
\[\tau_{m_2}(1-y_j(w^T\phi(x_j)+ b))+\xi_j-\epsilon_{m2}=0,\] which gives
\[ b= y_j-\sum_{i=1}^{l} \bigg(\alpha_{i}-\tau_1\alpha^{(1)}_{i},..~,-\tau_{k-1}\alpha^{(k-1)}_{i})k(x_i,x_j)\] \[ \hspace{-10mm}- y_j\frac{\epsilon_{m2}-\epsilon_{m1}}{(\tau_{m2}-\tau_{m1})} \bigg).\]

In practice, we compute all possible values of bias $b$ and take their average as final value of $b$.

\section{Properties of the $k$-PL-SVM model} 
In this section, we shall describe some interesting properties of the proposed $k$-PL-SVM models.
\subsection{Connection to existing SVM models}
The proposed $k$-PL-SVM model is a general SVM model. We shall show that there are several existing popular SVM models which can be realized as the particular cases of our $k$-PL-SVM model.
\begin{enumerate}
	\item [(a)] \textbf{C-SVM:-} The underlying Hinge loss function used in the $C$-SVM model is the particular case of our $k$-piece-wise linear loss function with $\tau_m =0$ and $\epsilon_{m}= 0$, for $m=1,2,..,k-1$. Also, if we consider the $\tau_m =0$ and $\epsilon_{m}= 0$, for $m=1,2,..,k$, in the optimization problem (\ref{pllsvm}) of proposed $k$-PL-SVM model, it becomes equivalent to the optimization problem of $C$-SVM model.
	
	\item [(b)] \textbf{Pin-SVM:-} The pinball loss function is also equivalent to $k$-piece-wise linear loss function with  $\tau_{1} =\tau$ for any $ -1 \leq \tau \leq 1$ , $\tau_{m_1} =0$ for $m_1=2,..,k-1$  and $\epsilon_{m_2}= 0$, for $m_2=1,..,k-1$ . At these value of parameters, the optimization problem (\ref{pllsvm}) of proposed $k$-PL-SVM model,  becomes equivalent to the optimization problem of Pin-SVM model.
	
	\item[(c)] \textbf {LS-SVM }:-  As we increase the value of $k$ in $k$-piece-wise linear loss function, it involves more segmented line. As $k \to \infty$, the $k$-piece-wise linear loss function becomes smooth and  the Least squares loss function becomes its particular case. Therefore, for very large value of $k$, the optimization problem (\ref{pllsvm}) of proposed $k$-PL-SVM model is equivalent to solving the optimization problem of LS-SVM model.
	
\end{enumerate}

\subsection{Geometrical Interpretation of $k$-PL-SVM model}
The $k$-piece-wise-linear loss functions form a family of convex and robust loss functions that exist between the Hinge loss function and Least squares loss function. In the $k$-PL-SVM model, the data points have been assigned the empirical risk according to their location. The $k$-PL-SVM model partition the feature space in different zones using  hyperplanes parallel to the separating hyperplanes $w^T\phi(x)+b =0$.  It assigns the empirical risk to a data point according to the zone it lies in.

\begin{figure}[H]
	\centering
	\includegraphics[width=1.0\linewidth]{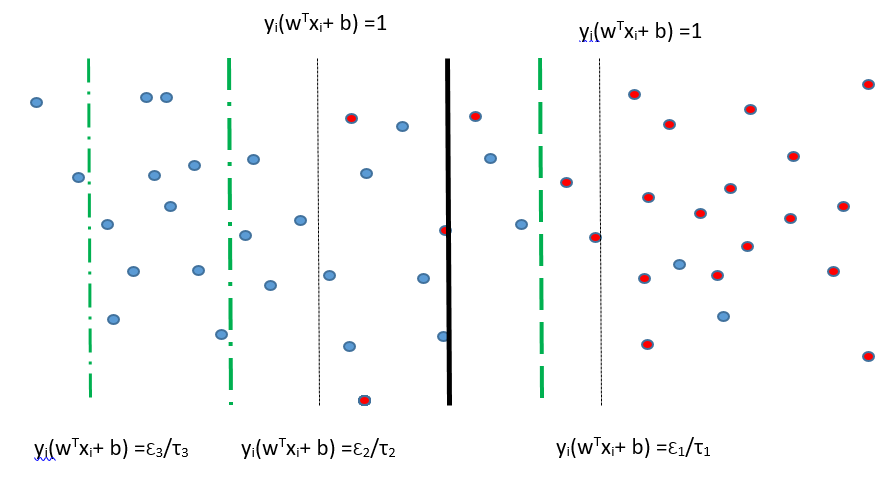}
	\caption{}
	\label{fig:geom}
\end{figure}

Let us suppose that for the given training set $T$, the $3$-PL-SVM model learns a suitable $3$-piece-wise linear loss function with  parameters $\tau_1, \epsilon_{1},\tau_{2}, \epsilon_{2},\tau_{3}$ and $\epsilon_{3}$, then similar to the C-SVM model, we have attempted to obtain the geometrical interpretation of our $3$-PL-SVM model in Figure \ref{fig:geom}. We have represented the data points from label $1$ with blue and label $-1$ with red color respectively.  We have shown the one possible division of the feature space in four different zones for data points with label $1$ in Figure \ref{fig:geom} by $3$-PL-SVM model. The 3-PL-SVM model assigns the empirical risk to the data points lying in these different zones using the different expression  which are in the form $-\tau_{k}(1-y_i(w^Tx_i+b)+ \epsilon_{k}) $ or $(1-y_i(w^Tx_i+b)$. A similar symmetric interpretation can be obtained for the data points with the label $-1$.

For $\tau_1, ...,\tau_{k-1} \geq 0 $ , we can easily show that the $k$-PL-SVM model minimizes the scatter of data points along the resulting decision function $w^T\phi(x)+b =0$.

It should be noted that it is not necessary that the $k$- PL-SVM model always learn a $k$-piece-wise linear loss function which contains $k$ different segment lines. It may learn values of its parameters $\tau_1, \epsilon_{1},....,\tau_{k-1}$ and $\epsilon_{k-1}$ from the given data such that the resulting loss function involves only $m$ segmented line where $m \leq k$. 

\subsection { Some properties of $k$-piece-wise linear loss function}
Now we shall evaluate the underlying $k$-piece-wise linear loss functions (\ref{pllos}) used in the proposed $k$-PL-SVM model using the existing literature of loss functions for classification problem. According to the study done in \cite{bartlett2006convexity} and \cite{pinsvmpath}, a typical classification loss function $L$ should have the  following four  properties.
\begin{itemize}
	\item[(a)]$L(u)$ should be Lipschitz for a given constant.
	\item [(b)] $L(u)$ should be convex.
	\item [(c)] $\frac{\partial{L(u)}}{\partial{u}} {|}_{u=1} >0 $.
	\item  [(d)] $L(u) \geq 0$ for any $u \in \mathbb{R}$. 
\end{itemize}
A  classification loss function which satisfies these four properties enjoys many  nice properties like Bayes consistency and classification calibration. It is not hard to realize that the existing Hinge loss function and pinball loss function with $\tau \geq 0$ satisfies these four properties.

The proposed $k$-piece-wise linear loss functions (\ref{pllos}) are Lipschitz and convex functions.  For $ \frac{\epsilon_{i}}{\tau_{i}} \neq 1, ~\forall ~i  =~1,2,..k-1$, we can easily obtain that the $k$-piece-wise linear loss functions  $L^k(u)$ satisfies $\frac{\partial{L^k(u)}}{\partial{u}} {|}_{u=1} >0 $. 

Our $k$-piece-wise linear loss functions can also take negative values for some values of its parameters. But, we can show that the $k$-piece-wise linear loss functions  which satisfy  $\frac{\epsilon_{i}\tau_{j}-\epsilon_{j}\tau_{i}}{\tau_{j}-\tau_{i}} \geq 0$ and $\frac{\epsilon_{j}}{1+\tau_{j}} \geq 0~, \forall i,j = 1, 2,..k-1$,  can always take non-negative values. Therefore, we can claim that similar to the Hinge loss function, the $k$-piece-wise linear loss functions with $\frac{\epsilon_{i}\tau_{j}-\epsilon_{j}\tau_{i}}{\tau_{j}-\tau_{i}} \geq 0$, $\frac{\epsilon_{j}}{1+\tau_{j}} \geq 0~$ and $  \frac{\epsilon_{i}}{\tau_{i}} \neq 1 ~\forall~, i,j = 1,2,..k-1$ enjoys the nice properties like Bayes consistency and classification calibration.   

Apart from this, similar to the Hinge loss function and the pinball loss function,  the proposed $k$-piece-wise linear loss functions are robust for finite values of $k$.  The influence function of the proposed family of $k$-piece-wise linear loss function can be shown to be bounded in the interval $[t_1,t_2]$, where $ t_1= min (1,-\tau_1,...,-\tau_{k-1})$ and $ t_1= max (1,-\tau_1,...,-\tau_{k-1})$. It means that any outlier data point can effect the resulting decision function up to a certain constant extent.
\section{Experimental Results}
In this section, we shall present the numerical results obtained by the extensive set of experiments and show the efficacy of the proposed $k$-PL-SVM model. We have compared the performance of the proposed 2-PL-SVM  and 3-PL SVM  with the C-SVM, LS-SVM and Pin-SVM models on benchmark datasets and shown that the proposed models own better generalization ability than existing SVM models.
\begin{table}
	\caption{Dataset Description}	
	\centering{\scriptsize
		\begin{tabular}{|c|l|l|l|}\hline
			Dataset	No. & 	Dataset & Size &  Training points  \\\hline 
			1	& Monk 1   &  556$\times$ 7    & 124                       \\
			2	 & Monk 2	&  601 $\times$7     &  169                    \\
			3	 & Monk 3 & 554 $\times$ 7    &     122                   \\
			4	 & Spect &    267 $\times$ 22    &    80                      \\ \hline
			5	& Haberman &  306 $\times$ 4  &   150                    \\  
			6	& Heart Statlog & 270 $\times$ 14   &    150  \\
			7  &Ionosphere &  351 $\times$ 34     &  200               \\
			8	  & Pima Indian &  768 $\times$ 9      &  300                \\
			9 & WDBC   & 569 $\times$ 30 & 400  \\ 
			10   & Echocardiogram & 131 $\times$ 10 &    80                \\ 
			11  & Australian  & 690  $\times$ 15   &  400                       \\ 
			12 & Bupa Liver &   345 $\times$ 7 &    250                \\
			13  & Votes & 435 $\times$ 17 &    200               \\ 
			14   & Diabetes & 768 $\times$ 9 & 500 \\ 
			15    & Fertility D. & 100 $\times$ 10  &   50                \\
			16     & Sonar & 208 $\times$ 61  & 100 \\   	        			 
			17     & Ecoil  & 327 $\times$ 8 & 200 \\ 
			18     & Plrx & 182 $\times$ 13 & 100 \\  
			19	 & Spambase &  4601 $\times$ 57 & 1500 \\ \hline
	\end{tabular}}
	
	\label{table1}
\end{table}
\begingroup
\setlength{\tabcolsep}{7pt} 
\renewcommand{\arraystretch}{1.2}
\begin{table}
		\caption{Numerical results with linear kernel}
	{\fontsize{5.3}{5.3} 	\selectfont 
		\begin{tabular}{|p{0.1cm}|c|c|c|c|c|}
			\hline
			Data                       & SVM       & LS-SVM   & Pin- SVM      & 2-PL SVM           & 3-PL-SVM                 \\ 
			\multirow{3}{*}{set}        & Acc.  & Acc.   & Acc.        & Acc.            & Acc.                   \\ 
			&  Time (s)     & Time (s)    & Time (s)        & Time (s)              & Time (s)                    \\ 
			No.	& $(C_0)$  & $(C_0)$ & $(C_0, \tau_1)$    & $(C_0, \tau_1, \epsilon_1)$      & $(C_0, \tau_1, \tau_2, \epsilon_1,\epsilon_{2})$    \\ \hline
			
			\multirow{3}{*}{1}        & 67.593    & 66.204   & 67.593        & 67.593             & 70.139                   \\ 
			& 0.055     & 0.051    & 0.052         & 0.054              & 0.067                    \\ 
			& (0.0625)  & (0.2500) & (0.0625,0)    & (0.0625,0,-5)      & (0.0625,1,-0.6,1.5,1)    \\ \hline
			\multirow{3}{*}{2}        & 67.13     & 67.13    & 67.13         & 67.13              & 67.361                   \\ 
			& 0.076     & 0.074    & 0.099         & 0.107              & 0.133                    \\ 
			& (0.0078)  & (0.0078) & (0.0078,-0.6) & (0.0078,-0.8,-5)   & (0.0078,-0.8,-1,-3,-3)   \\ \hline
			\multirow{3}{*}{3}        & 82.639    & 81.481   & 82.639        & 85.417             & 88.889                   \\ 
			& 0.051     & 0.050    & 0.057         & 0.058              & 0.065                    \\ 
			& (0.1250)  & (8.0000) & (0.1250,-0.6) & (0.1250,-0.4,0.5)  & (0.1250,-0.4,1,0.5,-3.5) \\ \hline
			\multirow{3}{*}{4}         & 76.471    & 74.332   & 83.957        & 83.957             & 83.957                   \\ 
			& 0.017     & 0.017    & 0.021         & 0.022              & 0.024                    \\ 
			& (2.0000)  & (0.0156) & (2.0000,-0.8) & (2.0000,-0.8,-0.5) & (2.0000,-0.8,-1,-0.5,-5) \\ \hline
			\multirow{3}{*}{5}      & 73.077    & 73.077   & 75.641        & 75.641             & 76.923                   \\ 
			& 0.036     & 0.035    & 0.059         & 0.057              & 0.054                    \\ 
			& (0.0078)  & (0.0078) & (0.0078,-0.8) & (0.0078,-0.8,0)    & (0.0078,1,-1,2,-5)       \\ \hline
			\multirow{3}{*}{6} & 85        & 84.167   & 85.833        & 85.833             & 87.5                     \\ 
			& 0.033     & 0.032    & 1.345         & 1.390              & 0.059                    \\ 
			& (0.2500)  & (0.0078) & (0.2500,-1)   & (0.2500,-1,-5)     & (0.2500,-0.4,0.2,1,-5)   \\ \hline
			\multirow{3}{*}{7}    & 92.715    & 93.377   & 92.715        & 92.715             & 92.715                   \\ 
			& 0.058     & 0.056    & 0.058         & 0.061              & 0.058                    \\ 
			& (2.0000)  & (4.0000) & (2.0000,0)    & (2.0000,0,-5)      & (2.0000,0,-1,-5,-5)      \\ \hline
			\multirow{3}{*}{8}   & 80.128    & 80.128   & 80.128        & 80.128             & 80.128                   \\ 
			& 0.177     & 0.172    & 0.176         & 0.189              & 0.175                    \\ 
			& (1.0000)  & (0.0625) & (1.0000,0)    & (1.0000,0,-5)      & (1.0000,0,-1,-5,-5)      \\ \hline
			\multirow{3}{*}{9}          & 98.225    & 97.041   & 98.225        & 98.817             & 98.817                   \\ 
			& 0.187     & 0.179    & 0.186         & 0.293              & 0.471                    \\ 
			& (0.0313)  & (0.0078) & (0.0313,0)    & (0.0313,0.2,0.5)   & (0.0313,0.2,-1,0.5,-5)   \\ \hline
			\multirow{3}{*}{10}    & 86.275    & 90.196   & 86.275        & 92.157             & 94.118                   \\ 
			& 0.009     & 0.008    & 0.408         & 0.019              & 0.013                    \\ 
			& (0.2500)  & (0.1250) & (0.2500,-1)   & (0.2500,-0.8,-4.5) & (0.2500,1,-0.2,2,0.5)    \\ \hline
			\multirow{3}{*}{11}    & 84.828    & 85.862   & 85.862        & 87.241             & 87.586                   \\ 
			& 0.223     & 0.207    & 11.870        & 11.890             & 1.388                    \\ 
			& (0.5000)  & (0.0156) & (0.5000,-1)   & (0.5000,-1,-0.5)   & (0.5000,-0.8,-0.8,4,-2)  \\ \hline
			\multirow{3}{*}{12}                     & 72.632    & 71.579   & 72.632        & 73.684             & 75.789                   \\ 
			& 0.069     & 0.065    & 0.068         & 0.104              & 0.148                    \\ 
			& (64.0000) & (1.0000) & (64.0000,0)   & (64.0000,-0.2,-5)  & (64.0000,-0.2,0.2,-1,-5) \\ \hline
			\multirow{3}{*}{13}         & 95.319    & 94.894   & 95.319        & 95.319             & 95.319                   \\ 
			& 0.069     & 0.066    & 0.069         & 0.072              & 0.069                    \\ 
			& (0.1250)  & (0.0156) & (0.1250,0)    & (0.1250,0,-5)      & (0.1250,0,-1,-5,-5)      \\ \hline
			\multirow{3}{*}{14}      & 81.716    & 81.716   & 81.716        & 81.716             & 82.836                   \\ 
			& 0.302     & 0.286    & 0.601         & 0.308              & 0.912                    \\ 
			& (1.0000)  & (0.0625) & (1.0000,-0.2) & (1.0000,0,-5)      & (1.0000,-0.2,0.8,0.5,-5) \\ \hline
			\multirow{3}{*}{15}    & 94        & 94       & 94            & 94                 & 94                       \\ 
			& 0.005     & 0.004    & 0.007         & 0.008              & 0.011                    \\ 
			& (0.0078)  & (0.0078) & (0.0078,-0.2) & (0.0078,-0.8,-5)   & (0.0078,-0.8,-1,-5,-5)   \\ \hline
			\multirow{3}{*}{16}         & 75.926    & 73.148   & 75.926        & 75.926             & 75.926                   \\ 
			& 0.019     & 0.017    & 0.018         & 0.019              & 0.019                    \\ 
			& (0.0313)  & (0.0156) & (0.0313,0)    & (0.0313,0,-5)      & (0.0313,0,-1,-5,-5)      \\ \hline
			\multirow{3}{*}{17}         & 94.488    & 81.89    & 95.276        & 95.276             & 95.276                   \\ 
			& 0.052     & 0.049    & 0.073         & 0.073              & 0.090                    \\ 
			& (0.0078)  & (0.5000) & (0.0078,-0.2) & (0.0078,-0.2,0)    & (0.0078,-0.2,-1,0.5,0.5) \\ \hline
			\multirow{3}{*}{18}          & 67.073    & 67.073   & 67.073        & 69.512             & 70.732                   \\ 
			& 0.015     & 0.014    & 0.655         & 0.642              & 0.022                    \\ 
			& (0.0078)  & (0.0078) & (0.0078,-1)   & (0.0078,-1,-2)     & (0.0078,1,0.8,2,-4)      \\ \hline
			
			\multirow{3}{*}{19}          & 88.562    & 83.034   & 88.562        & 88.562             & 89.768                   \\ 
			& 7.001     & 6.739    & 7.047         & 6.984              & 9.808                    \\ 
			& (8.0000)  & (2.0000) & (8.0000,0)   & (8.0000,0,-5)     & (8.0000,1,-0.6,2,-0.5)      \\ \hline
	\end{tabular}}
	\label{lintab}
\end{table}
\endgroup

 \begingroup
\setlength{\tabcolsep}{7pt} 
\renewcommand{\arraystretch}{1.1}
\begin{table}
		\caption{Numerical results with non-linear kernel}
	{\fontsize{5.3}{5.3} 	\selectfont 
		\begin{tabular}{|p{0.1cm}|c|c|c|c|c|}
			\hline
			Data                       & SVM       & LS-SVM   & Pin- SVM      & 2-PL SVM           & 3-PL-SVM                 \\ 
			\multirow{3}{*}{set}        & Acc.  & Acc.   & Acc.        & Acc.            & Acc.                   \\ 
			&  Time (s)     & Time (s)    & Time (s)        & Time (s)              & Time (s)                    \\ 
			No.	& $(q,C_0)$  & $(q,C_0)$ & $(q,C_0, \tau_1)$    & $(q,C_0, \tau_1, \epsilon_1)$      & $(q,C_0, \tau_1, \tau_2, \epsilon_1,\epsilon_{2})$    \\ \hline
			\multirow{3}{*}{1}  & 87.50      & 86.81       & 87.50           & 88.19             & 88.19                     \\ 
			& 0.154      & 0.157       & 0.155           & 0.167             & 0.180                     \\ 
			& (2,128)    & (2,128)     & (2,128,0)       & (2,128,0.2,1)     & (2,128,0.2,-1,1,-5)       \\ \hline
			\multirow{3}{*}{2}  & 86.11      & 87.04       & 86.11           & 86.34             & 87.27                     \\  
			& 0.238      & 0.241       & 0.248           & 0.251             & 0.255                     \\ 
			& (0.5,1)    & (1,8)       & (0.5,1,0)       & (0.5,1,0.4,0.5)   & (0.5,1,1,-0.4,2,0.5)      \\ \hline
			\multirow{3}{*}{3}  & 94.44      & 94.21       & 94.44           & 96.53             & 96.53                     \\ 
			& 0.151      & 0.154       & 0.150           & 0.163             & 0.180                     \\  
			& (4,16)     & (4,64)      & (4,16,0)        & (4,16,0.2,0.5)    & (4,16,0.2,-1,0.5,-5)      \\ \hline
			\multirow{3}{*}{6}  & 85.83      & 85.83       & 85.83           & 86.67             & 86.67                     \\ 
			& 0.093      & 0.092       & 0.093           & 0.135             & 0.183                     \\
			& (8,2)      & (4,0.03125) & (8,2,0)         & (8,2,-0.8,0.5)    & (8,2,-0.8,-1,-1.5,-4)     \\ \hline
			\multirow{3}{*}{9}  & 98.22      & 98.82       & 98.22           & 98.22             & 98.82                     \\ 
			& 0.543      & 0.534       & 0.544           & 0.539             & 1.013                     \\ 
			& (4,0.5)    & (4,8)       & (4,0.5,0)       & (4,0.5,0,-5)      & (4,0.5,0.2,-0.6,1.5,1)    \\ \hline
			\multirow{3}{*}{12} & 85.86      & 87.24       & 87.59           & 87.59             & 87.59                     \\ 
			& 0.663      & 0.634       & 0.949           & 0.961             & 1.277                     \\ 
			& (2,0.0625) & (4,0.0625)  & (2,0.0625,-0.8) & (2,0.0625,-0.8,0) & (2,0.0625,-0.8,-0.8,5,-5) \\ \hline
			\multirow{3}{*}{13} & 74.74      & 75.79       & 75.79           & 76.84             & 77.90                     \\ 
			& 0.129      & 0.126       & 0.168           & 0.166             & 0.224                     \\ 
			& (4,128)    & (4,128)     & (4,128,0.2)     & (4,128,0.2,0.5)   & (4,128,0.4,0.2,0.5,0.5)   \\ \hline
			\multirow{3}{*}{15} & 82.46      & 82.09       & 82.46           & 82.46             & 83.96                     \\ 
			& 0.561      & 0.558       & 0.563           & 0.566             & 1.128                     \\ 
			& (8,64)     & (8,8)       & (8,64,0)        & (8,64,0,-5)       & (8,64,-0.2,0.2,0.5,-0.5)  \\ \hline
	\end{tabular}}
	\label{nonlintab}
\end{table}
\endgroup

As we increase the value of $k$ in $k$-PL-SVM model, it becomes more adaptive and powerful. But, the optimization problem (\ref{pllsvm}) of $k$-PL-SVM model involves $k$ linear constraints which increases its solution time as value of $k$ increases. We could have solved the optimization problem (\ref{op_PLSVM}) of the proposed $k$-PL-SVM model using the stochastic gradient descent methods \cite{bottou2010large} to get rid of this problem. But, we need the accurate solution of our proposed $k$-PL-SVM model for studying its characteristics efficiently. Therefore, we have preferred to solve the  dual problem (\ref{dual_plsvm}) for the $k$-PL-SVM  models for $k=$~2 and 3 using the quadprog function in the MATLAB (\url{in.mathworks.com}) with the `interior point-convex' algorithm efficiently. We have also solved the C-SVM model and Pin-SVM model with the quadprog function in the MATLAB with the `interior point-convex' algorithm. The LS-SVR model only requires the solution of  system of equations.

     Also, the $k$-PL-SVM model requires to tune the $2k-1$ parameters for the linear kernel and $2k$ parameter for the RBF kernel.  As the value of $k$ increases, we require to tune more number of parameters in proposed $k$-PL-SVM model but, this extra endeavor in parameter tuning results in significant improvement in the prediction ability.

      In our numerical experiments, we have used the direct grid search method to tune the parameters of 3-PL-SM, 2-PL-SVM, Pin-SVM  and  LS-SVM models. We could have used some meta heuristic methods for tunning the parameters of these SVM models. But, these algorithms are of the random nature and may cause the difficulty in unbiased comparisons of considered SVM models. Therefore, we have preferred to use the direct grid search method to tune the parameters of considered SVM models in same ranges.

       It should ne noted that our objective of numerical experiments is to establish the fact that the proposed $k$-PL-SVM  models are improvement over existing SVM models.
      
      The C-SVM and LS-SVM models involve two parameters $C_0$ and RBF parameter $q$. The Pin-SVM model contains one  more extra parameter $\tau$. The 2-PL SVM model requires to tune $\tau_1$ and $\epsilon_1$ apart from $C_0$ and $q$. The 3-PL-SVM model requires the tuning of two more extra parameters $\tau_2$ and $\epsilon_{2}$ than 2-PL-SVM model. For the linear kernel, we do not require the use and tuning of kernel parameter $q$.  We have used the RBF kernel of the form $exp(\frac{-||x-y||_2} {2q^2})$.

       We have tuned the values of parameters $C_0$ and $q$ from the set $\{ 2^ {-7},2^{-6},......,2^6,2^7\} \times \{ 2^ {-7},2^{-6},......,2^6,2^7\} $ using grid search method for the C-SVM model. After obtaining the suitable values of these parameters,  we have used them  in Pin-SVM, 2-PL-SVM model and 3-PL-SVM model. The reason behind considering the same value for the parameters $C_0$ and $q$ in C-SVM, Pin-SVM , 2-PL SVM and 3-PL-SVM model is that we want to exclude the effects of these parameter on their performance. It helps us to make the observation that how tunning of extra parameters in Pin-SVM, 2-PL SVM and 3-PL-SVM model effects the improvement in accuracy.  The parameter $\tau$ in Pin-SVM model is obtained from the set of $\{ -1,-0.8,...,0.8,1\} $ using grid search method. The parameters $\tau_{1}$ and $\epsilon_1$  of 2-PL-SVM model has been obtained from the set $\{ -1,-0.8,...,0.8,1\}  \times \{ -5,-4.5,...,4.5,5\} $ using grid search method.  The parameters $\tau_{1}$, $\epsilon_1$,$\tau_2$ and $\epsilon_{2}$ have been obtained from the set $\{ -1,-0.8,...,0.8,1\}  \times \{ -5,-4.5,...,4.5,5\}  \times \{ -1,-0.8,...,0.8,1\}  \times \{ -5,-4.5,...,4.5,5\} $ using grid search method. For the LS-SVM model, we have explicitly tuned the parameters $C_0$ and $q$ from the set $\{ 2^ {-7},2^{-6},......,2^6,2^7\} \times \{2^{-7},2^{-6},......,2^6,2^7\} $ using grid search method. After obtaining the value of $C_0$, we have computed the $C_i$ from (\ref{ci}) for all SVM models.
       
       We have performed all experiments in MATLAB 2018 (\url{in.mathworks.com}) environment on a Dell Xeon processor with 16 GB of RAM and Windows 10 operating system.
       We have considered the 19 benchmark datasets for our numerical experiments. We have listed and numbered these datasets with their dimensions in Table \ref{table1}. These datasets have been downloaded from UCI repository \cite{UCIbenchmark}.  For some datasets like Monk 1, Monk 2, Monk 3 and Spect , the training sets and testing sets were separately provided. For other datasets, we have arbitrarily separated the training and testing set in  Table \ref{table1}.  Further, we have normalized the training and testing set in $[-1,1]$ for all datasets.

      \begin{figure*}[h]
      	\centering
      	\includegraphics[width=0.9\linewidth, height=0.2\textheight]{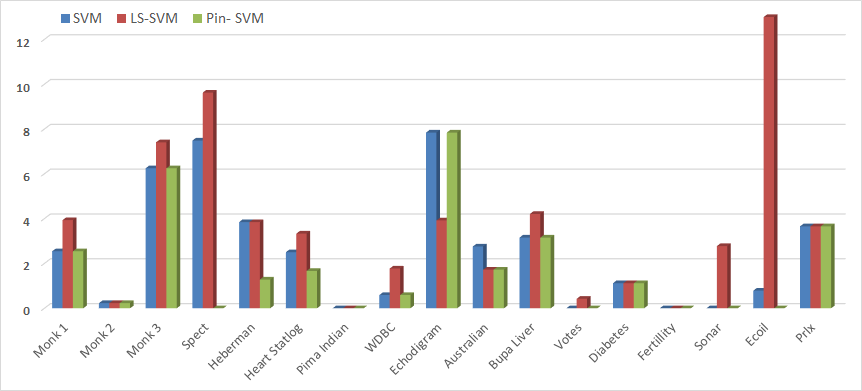}
      	\caption{Improvement in accuracy obtained by the proposed 3-PL-SVM model over existing $C$-SVM, LS-SVM and Pin-SVM models }
      	\label{impove3pl}
      \end{figure*}
      \begin{figure*}[h]
      	\centering
      	\subfloat[Monk 1]{\includegraphics[width=.25\linewidth, height = .15\textwidth ]{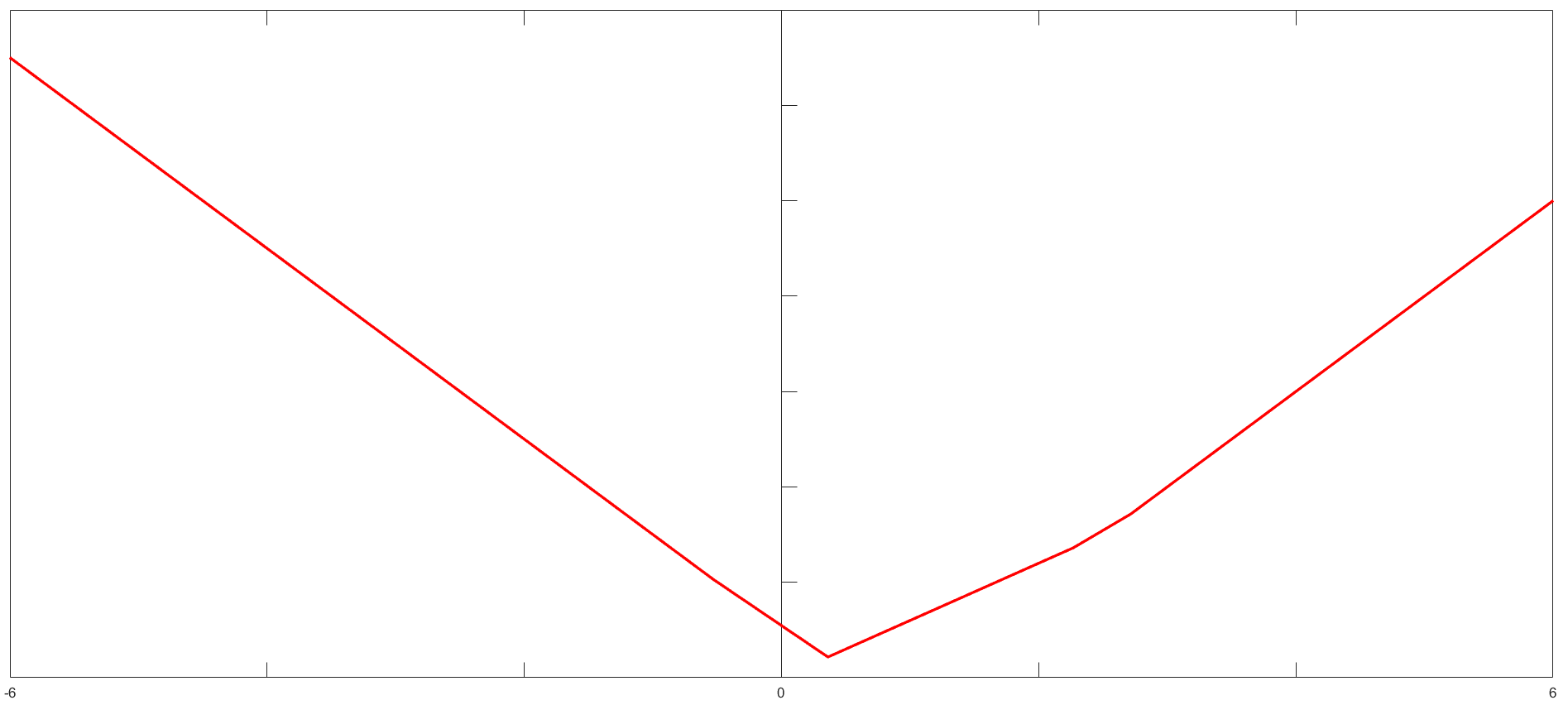}}
      	\subfloat[Monk 2]{\includegraphics[width=.25\linewidth, height = .15\textwidth ]{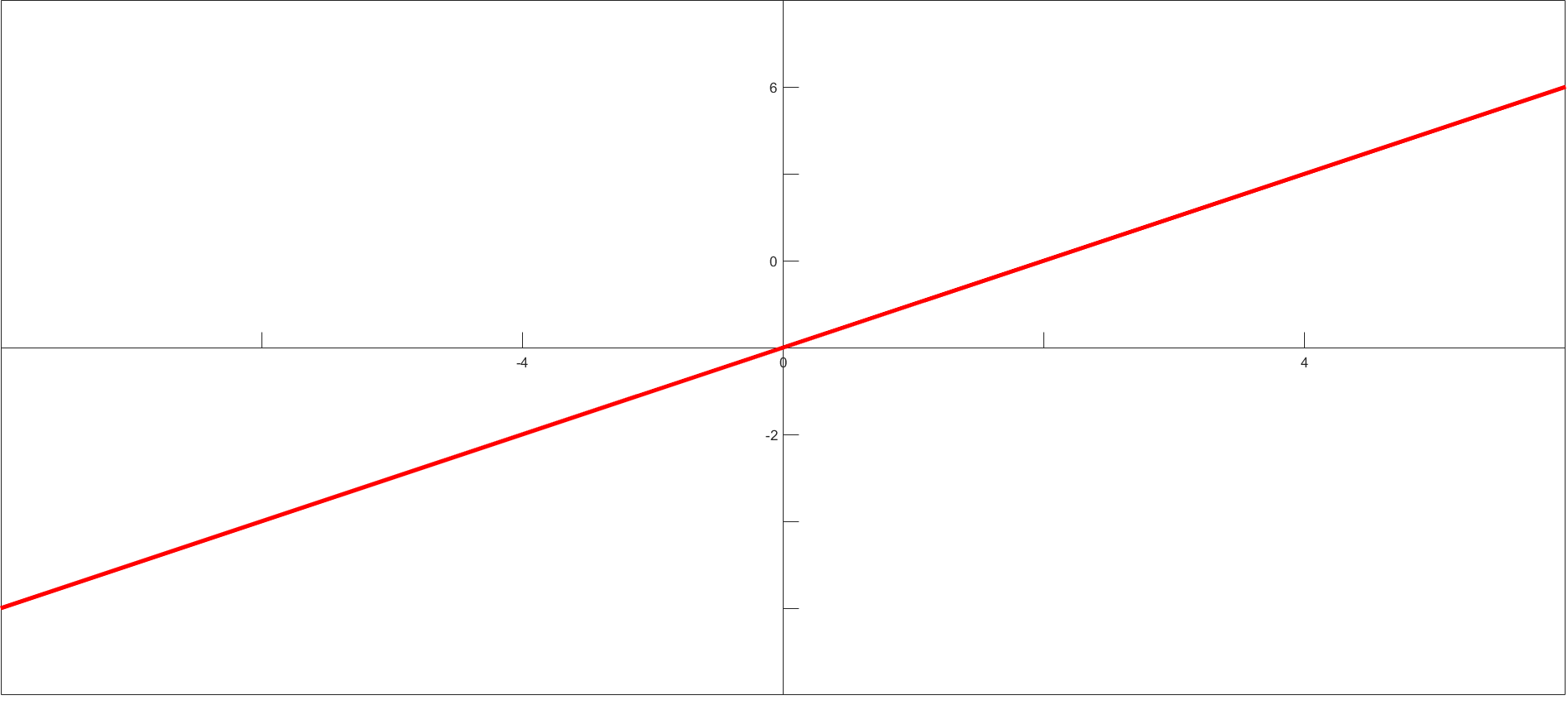}}
      	\subfloat[ Monk 3]{\includegraphics[width=.25\linewidth, height = .15\textwidth ]{ 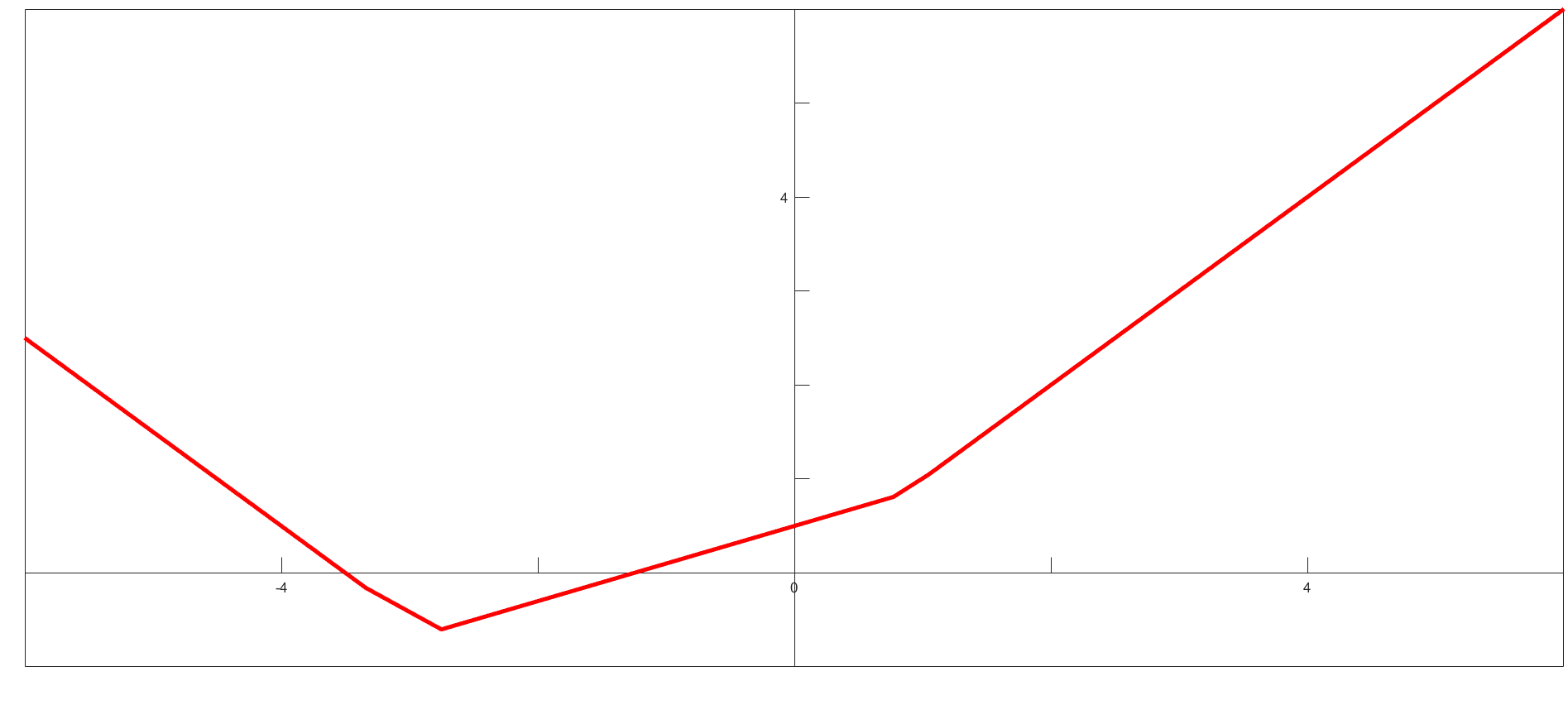}}
      	\subfloat[Heart]{\includegraphics[width=.25\linewidth, height = .15\textwidth ]{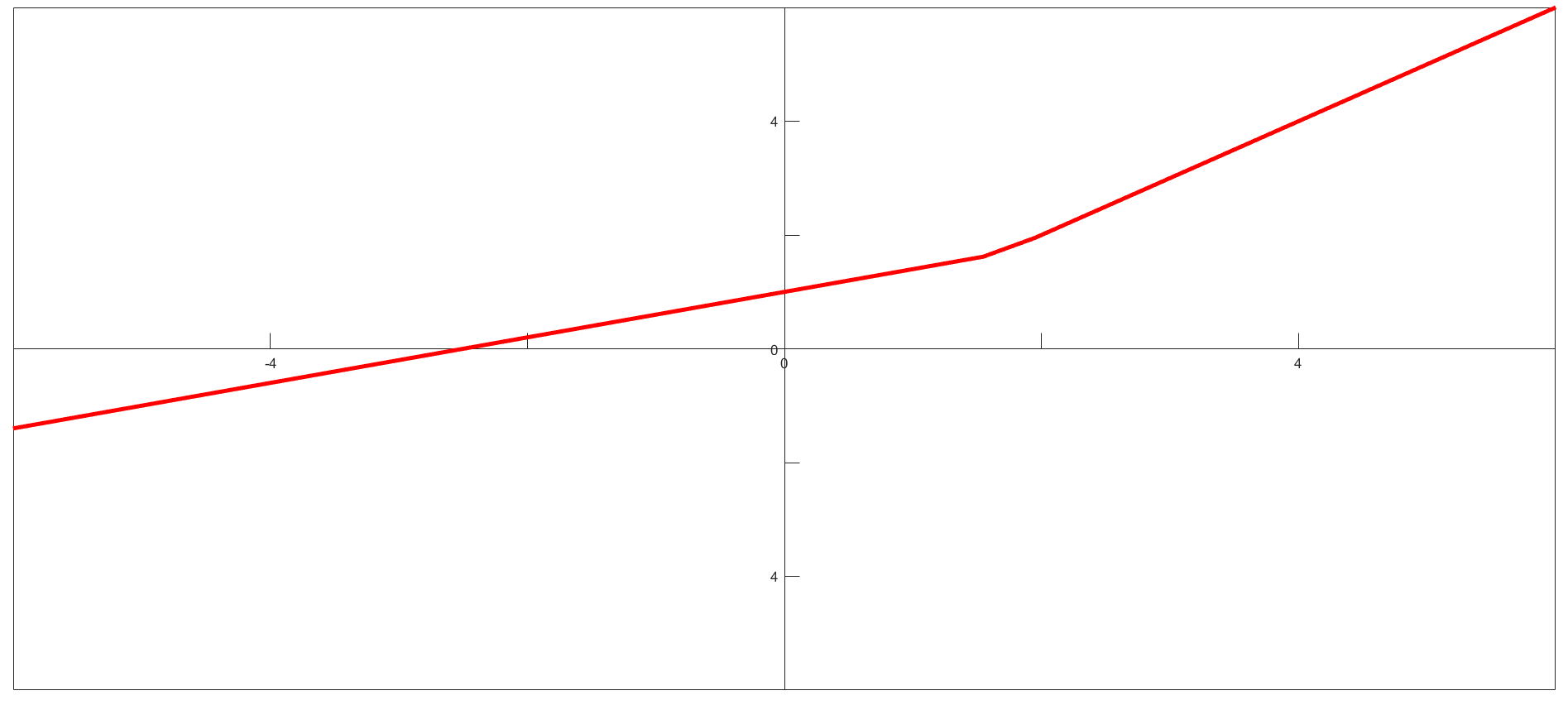}}\\
      	\subfloat[Haberman]{\includegraphics[width=.25\linewidth, height = .15\textwidth ]{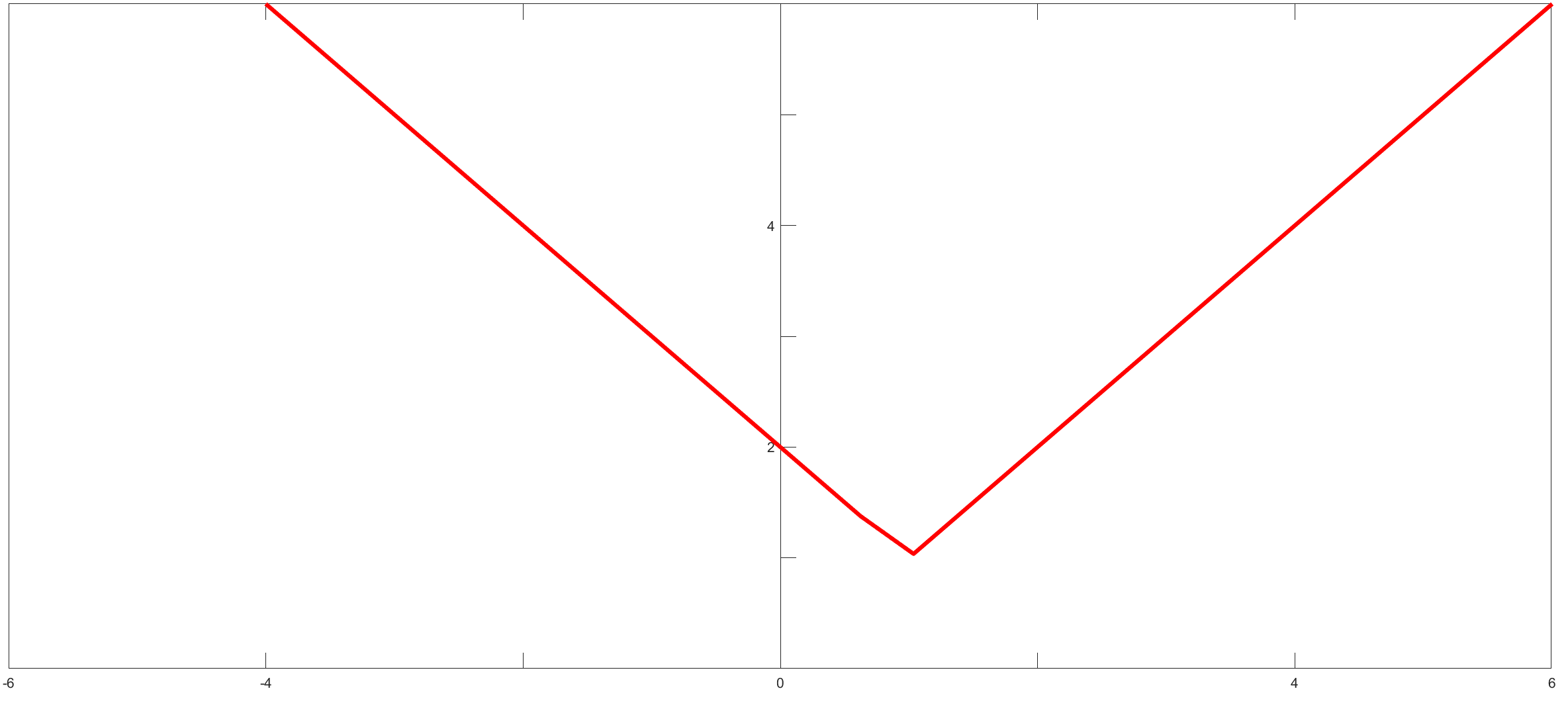}}
      	\subfloat[Echo]{\includegraphics[width=.25\linewidth, height = .15\textwidth ]{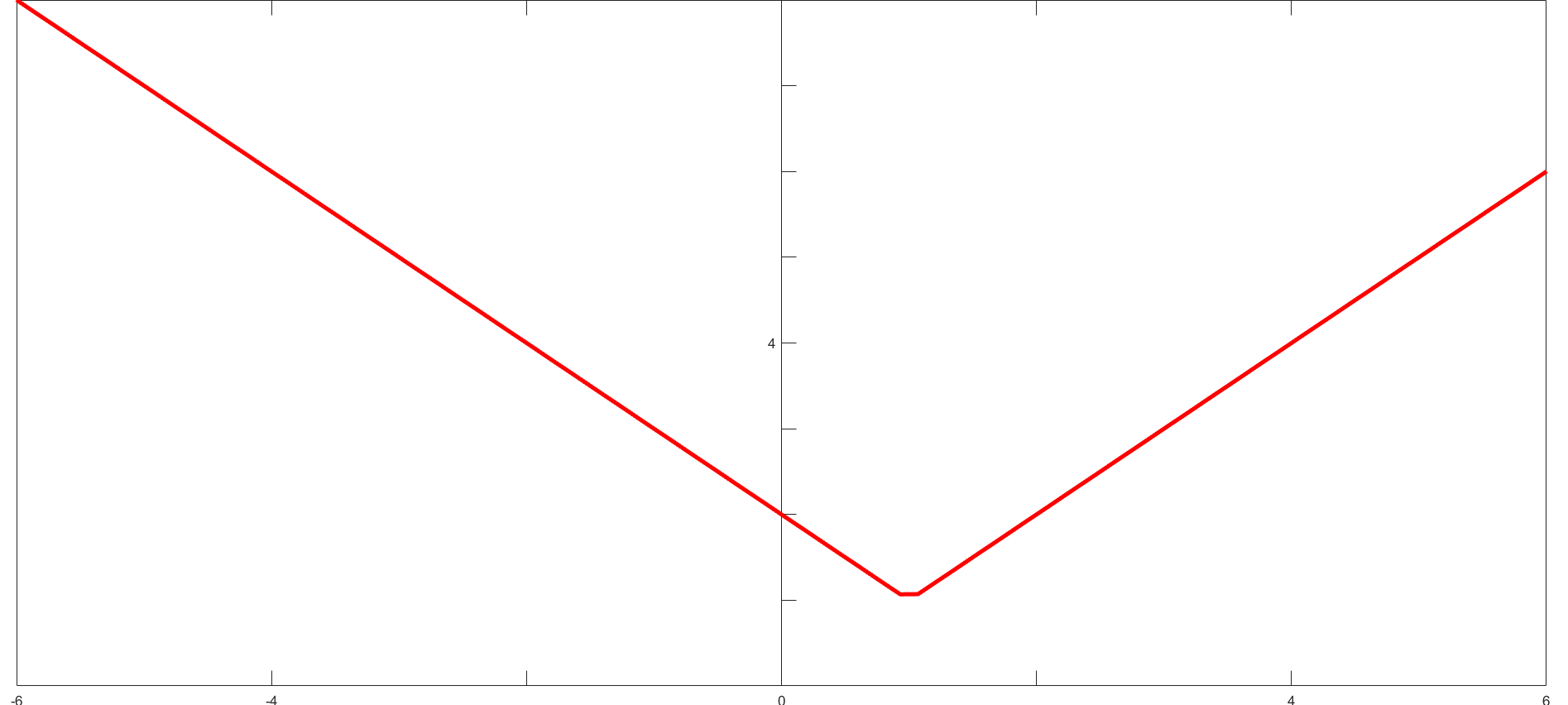}}
      	\subfloat[Bupa Liver]{\includegraphics[width=.25\linewidth, height = .15\textwidth ]{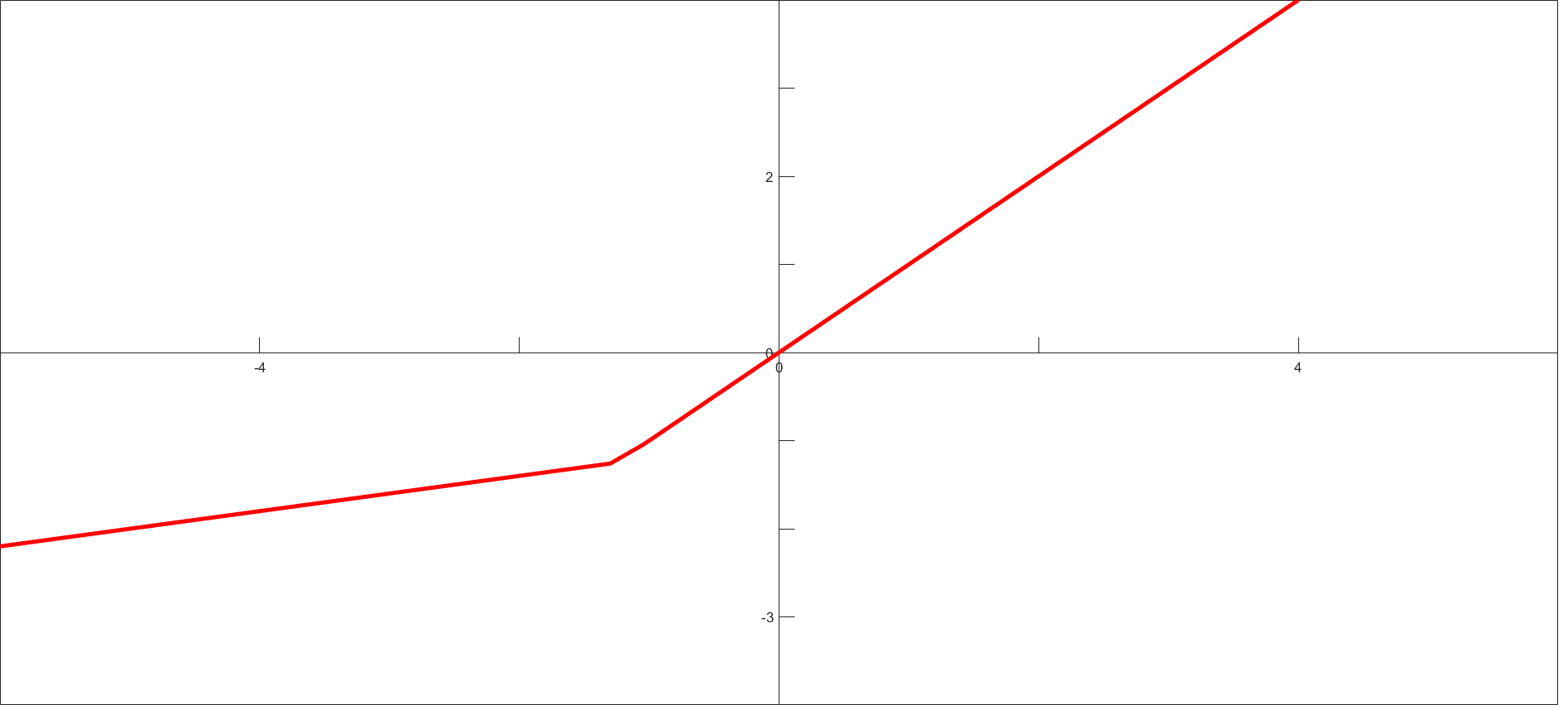}}
      	\subfloat[Diabetes]{\includegraphics[width=.25\linewidth, height = .15\textwidth ]{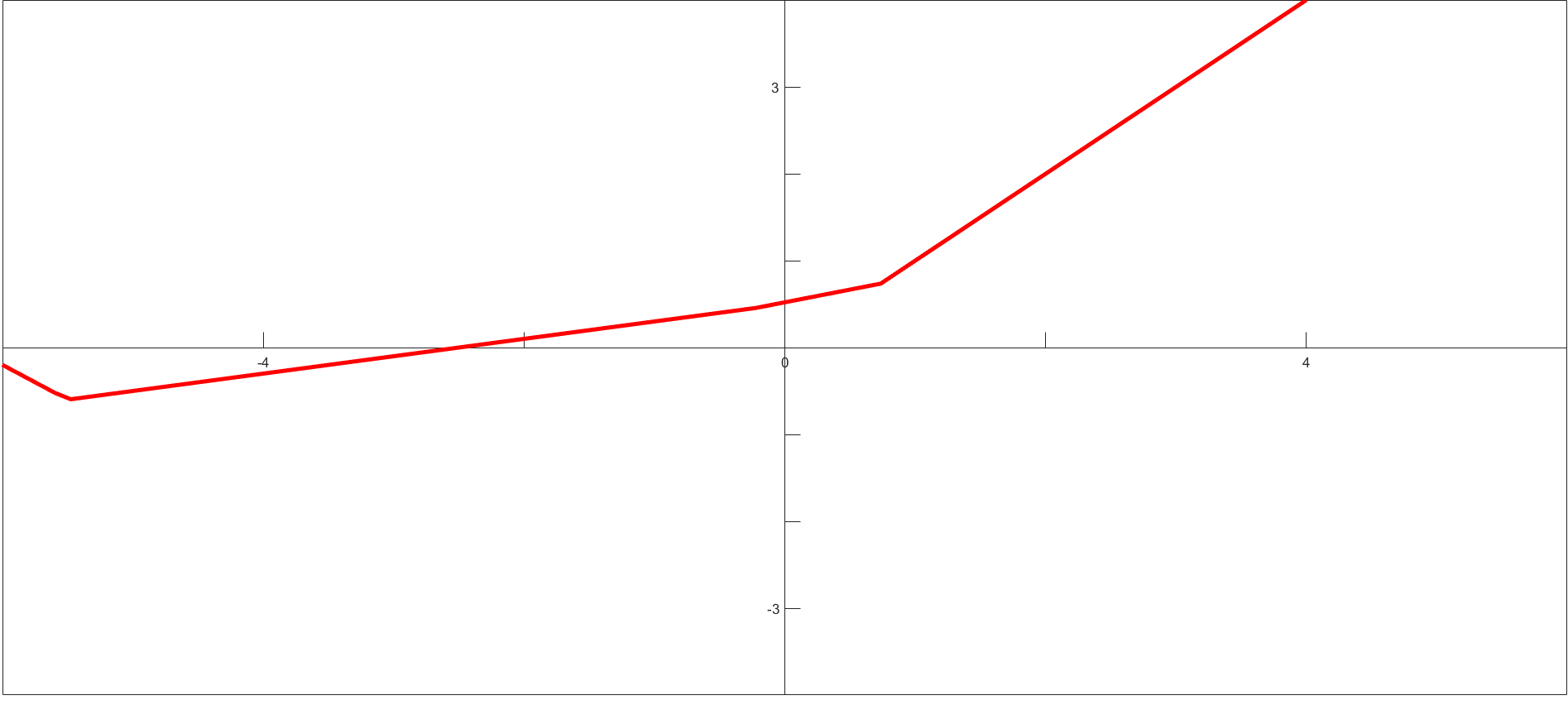}}
      	\caption{Loss functions learned by 3-PL-SVM model for different datasets.}
      	\label{losslearn}
      	
      \end{figure*}
      

We have listed and compared the performance of the   C-SVM, LS-SVM, Pin-SVM, 2-PL-SVM and 3-PL-SVM models in the Table \ref{lintab} for the linear kernel. We have also listed their tuned parameters and execution time. We can easily make the following observations form the numerical results listed in the Table \ref{lintab}.
\begin{itemize}
	\item  The accuracy obtained by SVM models in Table \ref{lintab} always follows the rule: C-SVM $\leq$ Pin-SVM $\leq$ 2-PL-SVM $\leq$ 3-PL-SVM. In most of datasets, as we move from C-SVM model to 3-PL-SVM model, there is a significant improvement in the accuracy. It means that the most of real world datasets finds the Hinge loss function rigid and need the use of adaptive loss functions for the better results. Also, it means that the tuning the extra parameters in $k$-PL-SVM models are useful.
	
	\item We can observe that the 3-PL-SVM model is powerful as it can obtain a direct improvement in accuracy over existing C-SVM, LS-SVM and Pin-SVM  model on 12 datasets. We have plotted this improvement obtained by 3-PL-SVM model in accuracy over C-SVM, LS-SVM  and Pin-SVM model  in the Figure \ref{impove3pl}. The use of 3-piece-wise linear loss function in 3-PL-SVM model makes it more efficient and adaptive. It enables the 3-PL-SVM model to learn the suitable  values of parameters $\tau_1$,$\tau_2$,$\epsilon_{1}$ and $\epsilon_{2}$. We have plotted the loss function learned by 3-PL-SVM in the Figure \ref{losslearn}. For different nature of datasets, the 3-PL-SVM has the capability to learn the suitable piece-wise linear loss function which is missing in the traditional SVM like C-SVM and LS-SVM model.
	
	\item The 2-PL-SVM model also improves significantly the C-SVM model on several datasets. The 3-PL-SVM model improves the 2-PL-SVM model on 11 datasets. As we increase the value of $k$ in $k$- PL-SVM model, its adaptive ability to learn the loss function according to the nature of dataset increases. It further results in the improvement of prediction ability.
	
	\item As we move from the C-SVM model to 3-PL-SVM model, the total execution time increases. It is because of the fact that we need to solve the QPP with more linear constraints.
\end{itemize}

We have also compared the performance of the  C-SVM, LS-SVM, Pin-SVM, 2-PL-SVM and 3-PL-SVM models for the RBF kernel in the Table \ref{nonlintab}. We can draw observations similar to the linear kernel case.

%
%
%
%

  \section{Conclusions}
  In this paper, we have developed a general and adaptive SVM model. For this, we have introduced a family of $k$-piece-wise linear loss functions in this paper. These loss functions are general, robust and convex. The resulting $k$-PL-SVM model can learn a suitable piece-wise linear loss function from the given data. The $k$-PL-SVM model is a general SVM model and the popular SVM models, like C-SVM, LS-SVM and Pin-SVM, are its particular cases.The $k$-PL-SVM model divides the feature space in different tube and assign the empirical risk to the data points according to its location. We have also shown that the $k$-PL-SVM model becomes Bayes consistent classification model if we impose certain constraint on the values of its parameter. We have carried an extensive set of experiments with the $k$-PL-SVM for $k=2 $ and $3$ and shown that the proposed SVM model is the improvement over existing SVM models.

We would like to check the performance of the $k$-PL-SVM model for $k \geq 4$ in future. We shall develop a suitable stochastic gradient descent method to efficiently approximate the solution of  the optimization problem of the $k$-PL-SVM model. It will enable  us to test the proposed $k$-PL-SVM model for large value of $k$. We have also planned to develop an appropriate meta heuristic algorithm for tuning the $2k$ parameters of the $k$-PL-SVM model in future. It will help us to reduce its total model selection time and also increase its adaptive capability. 

\section*{Acknowledgment}
I shall be very grateful to Prof. Suresh Chandra  for his valuable suggestions. His suggestions have significantly improve the quality of this  paper.
\bibliography{final}
\bibliographystyle{icml2021}

\appendix
%
%
%
%

\end{document}